\documentclass{llncs}
\usepackage{eepic,epsfig}
\usepackage{times}
\usepackage{helvet}
\usepackage{courier}
\usepackage{graphicx}
\usepackage{clrscode}

\title{Generic Global Constraints based on MDDs}

\author{
Peter Tiedemann \and Henrik Reif Andersen \and Rasmus Pagh}

\institute{IT University of Copenhagen\\
Rued Langgaards Vej 7\\DK-2300 Copenhagen S, Denmark\\
$\{petert,hra,pagh \}$@itu.dk
}

\begin{document}
\newcommand{\comment}[1]{}

\maketitle 

\begin{abstract}
  Constraint Programming (CP)\cite{CP-HANDBOOK} has been successfully
  applied to both constraint satisfaction and constraint optimization
  problems. A wide variety of specialized global constraints provide
  critical assistance in achieving a good model that can take
  advantage of the structure of the problem in the search for a
  solution. However, a key outstanding issue is the representation of
  'ad-hoc' constraints that do not have an inherent combinatorial
  nature, and hence are not modelled well using narrowly specialized
  global constraints. We attempt to address this issue by considering
  a hybrid of search and compilation. Specifically we suggest the use
  of Reduced Ordered Multi-Valued Decision Diagrams (ROMDDs) as the
  supporting data structure for a generic global constraint. We give
  an algorithm for maintaining generalized arc consistency (GAC) on
  this constraint that amortizes the cost of the GAC computation over
  a root-to-leaf path in the search tree without requiring
  asymptotically more space than used for the MDD. Furthermore we
  present an approach for incrementally maintaining the reduced
  property of the MDD during the search, and show how this can be used
  for providing domain entailment detection. Finally we discuss how to
  apply our approach to other similar data structures such as AOMDDs
  and Case DAGs. The technique used can be seen as an extension of the
  GAC algorithm for the regular language constraint on finite length
  input \cite{REGULAR-CONSTRAINT}.
\end{abstract}

\section{Introduction}
Constraint Programming (CP)\cite{CP-HANDBOOK} is a powerful technique
for specifying Constraint Satisfaction Problems (CSPs) based on
allowing a constraint programmer to model problems in terms of
high-level constraints. Using such \emph{global constraints} allows
easier specification of problems but also allows for faster solvers
that take advantage of the structure in the problem. 

The classical approach to CSP solving is to explore the search tree of
all possible assignments to the variables in a depth-first search
backtracking manner, guided by various heuristics, until a solution is
found or proven not to exist.  One of the most basic techniques for
reducing the number of search tree nodes explored is to perform
\emph{domain propagation} at each node. In order to get as much domain
propagation as possible we wish for each constraint to remove from the
variable domains all values that cannot participate in a solution to
that constraint.  This property is known as Generalized Arc
Consistency (GAC). 

It is only possible to achieve GAC for some types of global
constraints in practice, as a global constraint can model NP-hard
problems making it infeasible to achieve GAC. The use of global
constraints can significantly reduce the total number of constraints
in the model, which again improves domain propagation if GAC or other
powerful types of consistency can be enforced. However, in typical
CSPs there are many constraints that lie outside the domain of the
current global constraints. Such constraints are typically represented
as a conjunction of simple logical constraints or stored in tabular
form.  The former can potentially reduce the amount of domain
propagation, while the tabular constraints typically takes up too much
space for all but the most simple constraints, and hence it is also
computationally expensive to achieve GAC.

The aim of this paper is to introduce a new generic global constraint
type for constraints on finite domains based on the approach of
\emph{compiling} an explicit, but compressed, representation of the
solution space of as many constraints as possible. To this end we
suggest the use of Multi-Valued Decision Diagrams (MDDs). 

It is already known how to perform GAC in linear, or nearly linear
time in the size of the decision diagram for many types of decision
diagrams including MDDs \cite{BRYANT-BDD,AOMDD,VD-NOTE}.  However,
compact as decision diagrams may be, they are still of exponential
size in the number of variables in the worst case. In practice their
size is also the main concern, even when they do not exhibit worst
case behavior. Applying the static GAC algorithms at every step of the
search is therefore likely to cause an unacceptable overhead in many
cases. 

To avoid the overhead it is essential to avoid repeating computation
from scratch at each step and instead use an algorithm that amortizes
the cost of the GAC computation over a number of domain propagation
steps. In this paper we introduce such an algorithm using ideas from
\cite{REGULAR-CONSTRAINT}. The paper is structured as follows. In
Section \ref{subsec:related} we discuss in some detail how are our
contributions are related to those contained in
\cite{REGULAR-CONSTRAINT} as well as briefly describe other related
results. In Section \ref{sec:mdd-search} we describe the functionality
required from the global constraint we are presenting and the setup in
which it can be used. In Section \ref{sec:simple-VD} we describe how
the technique from \cite{REGULAR-CONSTRAINT} can be applied to a
simplified MDD. We then move on to show how we can extend that
approach to handle long edges in MDDs in Section
\ref{sec:long-edges}. In Section \ref{sec:reduction} we present
results on maintaining the reduced property of the constraint during a
search. Based on this we show how domain entailment can be detected in
Section \ref{subsec:domain_entailment}. Finally we discuss how our
approach can be applied to other data structures that are similar to
MDDs, such as AOMDDs and Case DAGs in Section \ref{sec:applications}.

\subsection{Related Work}\label{subsec:related}
The concept of compiling an explicit, but compact, representation of
the solution space of a set of constraints has previously been applied
to obtain backtrack-free configurators for many practical
configuration problems \cite{BDD-CONFIG}. In this case Binary Decision
Diagrams (BDDs) \cite{BRYANT-BDD} are used for representing the
solution space. However, it is well known that BDDs (and MDDs) are not
capable of efficiently representing prominent constraints such as the
AllDifferent constraint \cite{ALLDIFF}.

A regular language constraint for finite sequences of variables is
introduced in \cite{REGULAR-CONSTRAINT}. It uses a DFA to represent
the valid inputs where the input is limited to be of length $n$. Since
the constraint considers a finite number of inputs, these can be
mapped to $n$ variables, and GAC can be enforced according to the
domains of these variables. To this end the cycles in the DFA are
'unfolded' by taking advantage of the fact that the input is of a
finite length. The resulting data structure has size $O(nd_{max}q)$
where $q$ is the number of states in the original DFA and $d_{max}$
the size of the largest variable domain. A bounded incremental GAC
algorithm with complexity linear in the number of data structure
changes is also presented. Since this regular global constraint is
defined on a finite length input, it can be used as a generic
constraint. In fact, we note that there is a strong correspondence
between the unfolded DFA and a simplified MDD representing the same
constraint. However, there are some important theoretical and
practical reasons for using fully reduced MDDs when the goal is a
generic global constraint.  Below we summarize our contributions and
highlight the differences compared to using the regular constraint in
the role of a generic global constraint.

\begin{itemize}
\item Firstly, DFAs do not allow skipping inputs, even for states
  where the next input is irrelevant. Skipping input variables in this
  manner is part of the reduce steps for BDDs, and if it is used in
  MDDs it requires alterations to the GAC algorithm. We give a
  modified algorithm to handle this. In some cases allowing the
  decision diagram to skip variables can give a significant reduction
  in size. A very simple example is a constraint specifying that the
  value $v$ must occur at least once for one of the variables $x_1,
  \ldots, x_n$ each having the same domain of size $d$. In an MDD that
  does not allow skipped variables (such as an unfolded DFA) this
  requires $\Omega(n^2d^2)$ nodes compared to $O(nd)$ nodes if we
  allow skipped variables in the MDD.

\item Secondly, BDDs are normally kept reduced during operations on
  the BDDs.  This allows subsequent operations to run faster and also
  shows directly if the result is the constant true function. We
  present an approach that can dynamically reduce the MDD without
  resorting to scanning the entire live part of the data structure,
  and which also allows us to detect domain entailment
  \cite{vanhentenryck94design}. Domain entailment detection can be
  critical for the performance of a CSP solver, as it can save a
  potentially exponential number of exectutions of the GAC algorithm.
  The technique described in \cite{REGULAR-CONSTRAINT} does not
  provide dynamic reduction or domain entailment detection.

  The suggestion in \cite{REGULAR-CONSTRAINT} is to minimize the DFA
  only once at the beginning (thereby also minimizing the initial
  'unfolded' DFA) which would correspond to reducing the MDD prior to
  the search, and does not provide any form of entailment detection.
  The problem of efficient dynamic minimization is not discussed in
  \cite{REGULAR-CONSTRAINT} and would seem to require a technique
  similar to the one we present in this paper for obtaining dynamic
  reduction of the MDD constraint.
\item Thirdly, we cover how the GAC algorithm can be adapted to
  operate on other decision diagram style data structures such as
  AOMDDs \cite{AOMDD} and Case DAGs \cite{SICTUS-PROLOG-MANUAL}.
\end{itemize}

Finally, from a practical perspective it is not a good idea to first
construct a DFA and then unfold it. It is more efficient to use a BDD
package to construct an ROBDD directly, as efficient BDD packages
\cite{BUDDY,CUDD} with a focus on optimizing the construction phase
have already been developed driven by needs in formal verification
\cite{VLSI}. Specifically the use of BDDs for the construction gives
access to the extensive work done on variable ordering (see for
example \cite{FORCE,SCATTER,SIFTING}) for BDDs.  Once an ROBDD is
constructed it can then easily be converted into the desired MDD.

Another related result is \cite{ADHOC-GAC} in which it is discussed
how to maintain Generalized Arc Consistency in a generic global
constraint on binary variables based on a BDD. Their technique differs
from the straightforward DFS scanning technique by using shared
good/no-good recording and a simple cut-off technique to in some cases
reduce the amount of nodes visited in a scan. Their technique can be
adapted for non-binary variables, but the cut-off technique lose merit
if the MDD is to be reduced dynamically and becomes much less
efficient as the domains increase in size. Hence their techniques do
not apply when the intention is to collect sets of small constraints
into a few larger MDD constraints.

\subsection{Notation}
In this paper we consider a CSP problem $\textit{CP}(X,D,F)$, where $X
= \{ x_1, \ldots , x_n \}$ is the set of variables, $F$ the set of
constraints on the variables in $X$ and $D = \{ D_1, \ldots, D_n\}$ is
the multi-set of variable domains, such that the domain of a variable
$x_i$ is $D_i$. We use $d_i = |D_i|$ to denote the size of domains and
$d_{max} = \max \{ d_i \mid x_i \in X \}$ to denote the largest
domain. A \emph{single assignment} is a pair $(x_i,v)$ where $x_i \in
X$ and $v \in D_i$. The assignment $(x_i,v)$ is said to have support
in a constraint $F_k$, iff there exists a solution to $F_k$ where
$x_i$ is assigned $v$. If a single assignment ($x_i,v$) for which $v
\in D_i$ has support in a constraint $F_j$, $v$ is said to be in the
valid domain for $x_i$ respective to $F_j$ , denoted
$\textrm{VD}_i(F_j)$. If for all variables $x_i$ and a constraint
$F_j$ it is the case that $D_i = \textrm{VD}_i(F_j)$ then $F_j$ is
said to fulfill the property of \emph{Generalized Arc Consistency}
(GAC). A \emph{partial assignment} $\rho$ is a set of single
assignments to distinct variables, and a \emph{full assignment} is a
partial assignment that assigns values to all variables.

We now define an MDD constraint belonging to a specific CSP.

\begin{definition}[Ordered Multi-Valued Decision Diagram (OMDD)]
  An Ordered Multi-Valued Decision Diagram (OMDD or just MDD) for a
  CSP \textit{CP} is a layered Directed Acyclic MultiGraph $G(V,E)$
  with $n+1$ layers (some of which may be empty). Each node $u$ has a
  label $l(u) \in \{1, \ldots, n+1\}$ corresponding to the layer in
  which the node is placed, and each edge $e$ outgoing from layer $i$
  has a value label $v(e) \in D_i$.  Furthermore we use $s(e)$ and
  $d(e)$ to denote respectively the source and destination layer of
  each edge $e$ .

The following restrictions apply:
\begin{itemize}
\item There is exactly one node $u$ such that $l(u) = \min\{ l(q) \mid
  q \in V\}$ denoted \textit{root}.
\item There is exactly one node $u$ such that $l(u) = n+1$ denoted
\textit{terminal}.
\item For any node $u$, all outgoing edges from $u$ have distinct
labels.
\item All nodes except \textit{terminal} has at least one outgoing
  edge.
\item For all $e \in E$ it is the case that $s(e) < d(e)$.
\end{itemize}
\end{definition}

We will use $V_i = \{ u \in V \mid l(u) = i\}$ to denote the nodes of
layer $i$ and $E_i = \{ e \in E \mid s(e) = i\}$ to denote the set of
edges originating from layer $i$. Furthermore we define 
$$P_u = \{(p,v) \mid \exists e = (p,u) \in E : v(e) = v \}$$
and 
$$C_u = \{ (c,v) \mid \exists e = (u,c ) \in E : v(e) = v\}$$ 

That is, $P_u$ corresponds to the incoming edges to $u$, and $C_u$
corresponds to the outgoing edges of $u$. Given layers $i > j > k$, we
say that $i$ is a \emph{later} layer than $j$ and that $k$ is an
\emph{earlier} layer than $j$.

\begin{definition}[Solution to an MDD]
  A full assignment $\rho$ is a solution to a given MDD iff there
  exists a path $Q = (e_1,\ldots, e_j)$ from \textit{root} to
  \textit{terminal} such that for each $(x_i,v) \in \rho$ at least one
  of the following conditions hold:
\begin{itemize}
\item $\exists e \in Q$ such that $s(e) = i$ and $v(e) =
  v$
\item $l(root) > i$
\item $\exists e \in Q$ such that $s(e) < i < d(e)$.
\end{itemize}
\end{definition}

An edge $e$ such that $s(e) +1 < d(e)$, as in the third condition in
the above definition, is called a \emph{long edge} and is said to skip
layer $s(e) + 1$ to $d(e) -1$. It it is worth noting that one long
edge can represent many partial assignments.

\begin{definition}[Reduced OMDD]
  An MDD is called \emph{uniqueness reduced} iff for any two distinct
  nodes $u_1,u_2$ at any layer $i$ it is the case that $C_{u_1} \not =
  C_{u_2}$. If it is furthermore the case for all layers $i \in \{1,
  \ldots,n\}$ that no node $u_1$ in layer $i$ exists with $d_{i}$
  outgoing edges to the same node $u_2$, the MDD is said to be
  \emph{fully reduced}.
\end{definition}

\begin{figure}
\begin{center}
\includegraphics[height=6cm]{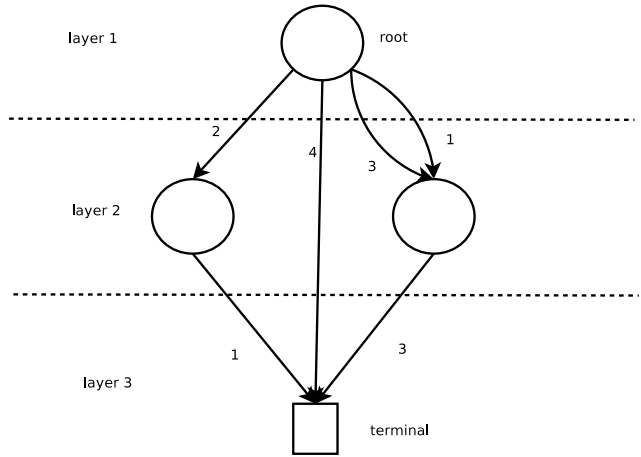}
\end{center}
\caption{The above figure shows an example MDD that is fully
  reduced. Assuming $D_1= \{ 1,2,3,4\}$ and $D_2 = \{ 1,2,3\}$, it
  represents the binary constraint with solutions \{ (1,3),(2,1),
  (3,3),(4,1),(4,2),(4,3) \}.}
\label{fig:example}
\end{figure}

The above definitions are just the straightforward extension of the
similar properties of BDDs\cite{BRYANT-BDD}. Fully reduced MDDs retain
the canonicity property of reduced BDDs, that is, for each ordering of
the variables there is exactly one fully reduced MDD for each Boolean
constraint on $n$ discrete domain variables. An example reduced MDD is
shown in Figure \ref{fig:example}.

\section{Searching with an MDD}\label{sec:mdd-search}
In this paper we consider a backtracking search for a solution to a
conjunction of constraints, at least one of which is an MDD. To
simplify the complexity analysis, we assume that the search branches
on the domain values of each variable in some specified order and that
full domain propagation takes place after each branching. The process
of branching and performing full domain propagation we will refer to
as a \emph{phase}. For our proposed global constraint we refer to
performing domain propagation on it as a single \emph{step}. As such,
one phase may contain many steps depending on how many iterations it
takes until none of the constraints are able to remove any further
domain values. In order to be useful in this type of CSP search, an
implementation of the MDD constraint needs to supply the following
functionality:

\begin{itemize}
\item \proc{Assign}($x,v$)
\item \proc{Remove}($(x_1,v_1),\ldots, (x_k,v_k)$)
\item \proc{Backtrack}()
\end{itemize}

The \proc{Assign} operation restricts the valid domain of $x$ to $v$
and is used to perform branchings. The \proc{Remove} operation
corresponds to domain restrictions occurring due to domain propagation
in the other constraints. The \proc{Backtrack} operations undoes the
last \proc{Assign} operation and all \proc{Remove} operations that has
occurred since, effectively backtracking one phase in the search
tree. For the implementation of \proc{Backtrack} we will simply push
data structure changes on a stack, so that they can be reversed easily
when a backtrack is requested. This very simple method ensures that a
backtrack can be performed in time linear in the number of data
structure changes made in the last phase. For all the dynamic data
structures considered in this paper the space used for this undo stack
will be asymptotically bounded by the time used over a root-to-leaf
path in the search tree. Furthermore, in all the cases studied in this
paper, \proc{Assign}($x_i,v$) is just as efficiently implemented as a
single call to \proc{Remove}($ \{ (x_i,v_j) \mid v_j \in D_i \setminus
\{ v \} \}$). Therefore we will only discuss the implementation of
\proc{Remove}.

\section{Calculating the change in valid domains}\label{sec:simple-VD}
In the following we will say that an edge $e$ is valid iff there
exists a path in the MDD, corresponding to a solution to the MDD under
all restrictions applied so far, such that $e$ is part of the path. We
furthermore say that a node is valid if it is \textit{root} or has at
least one valid incoming and outgoing edge. A crucial element in
computing the valid domains in an MDD is that of a \emph{supporting
  edge}. A valid edge $e \in E$ \emph{supports} a single assignment
$(x_i,v)$, if $s(e) = i$ and $v(e) = v$ or if $s(e) < i < d(e)$.  Note
that $v \in D_i$ iff there exists an edge $e$ supporting the
assignment $(x_i,v)$.  We say that a node $u$ supports a value $v$ if
there exists an edge $(u,c)$ supporting $(x_{l(u},v)$ for some $c$.

In this section we will show how the approach from
\cite{REGULAR-CONSTRAINT} can be used for maintaining the valid
domains of a simplified MDD. We will temporarily assume that the OMDD
we operate on is initially Uniqueness Reduced, but not fully reduced,
in fact we will assume that all outgoing nodes from a node in layer
$i$ lead to nodes in layer $i+1$ (and hence also that $l(root) = 1$).

\subsection{Tracking support}
Valid domains are maintained by tracking the loss of supporting edges
for each possible assignment as restrictions of the form $x_i \not =
v$ are applied. The application of a restriction $x_i \not = v$
immediatly invalidates all edges originating from layer $i$ with label
$v$. We need to compute what further supporting edges are invalidated
as a consequence and determine if any assignments have lost their
supporting edges.

We track the set of supporting edges by storing a set of sets $S$,
such that for every possible single assignment $(x_i,v)$ where $x_i
\in X$ and $v \in D_i$ there exists a set $s_{i,v} \in S$ containing
all the nodes and the corresponding edges that gives support to the
single assignment $(x_i,v)$. In maintaining $S$ we learn immediately
when a single assignment $(x_i,v)$ no longer has support, as $s_{i,v}$
will be empty. Note that the space needed for the support lists is
only $O(|E|)$. The approach to maintaining $S$ is very simple. To
apply a restriction $x_i \not = v$, we first invalidate all edges
supporting this assignment. We then check if this has made any nodes
invalid, and if so invalidate their edges, continuing recursively.

\subsection{Performing \proc{Remove}}
The above support tracking technique is implemented as the procedures
\proc{Remove} and \proc{RemoveEdge} shown in Figure
\ref{code:inc_remove}. The remove operation is performed by, for each
single assignment $(x_i,v)$ to be removed, visiting all nodes that
offer support for $(x_i,v)$. On each such node the update procedure
\proc{RemoveEdge} is used to remove the corresponding invalidated edge
while maintaining $S$ and the valid domains.

\subsection{Complexity}
The \proc{RemoveEdge} operation can easily be supported in $O(1)$ time
per call using $O(|E|)$ space using just linked lists for the support
sets and pointers from each edge to their corresponding support list
entry. 

Consider a path in the search tree implicitly represented by the
branching search. We will here and in the following describe the
complexity of the presented algorithms as the complexity over such a
root-to-leaf path in the search tree. Since each edge can only be
removed once during this part of the search we obtain the following
result:

\begin{lemma}\label{lemma:path-complexity}
Consider any root-to-leaf path in the search tree. Then the total
number of calls to \proc{RemoveEdge} is at most $O(|E|). $
\end{lemma}

\section{Skipping input variables} \label{sec:long-edges} We have so
far obtained a very efficient GAC algorithm for a simplified MDD data
structure by simply applying the technique from
\cite{REGULAR-CONSTRAINT} to our MDD constraint. However, the
algorithm described so far does not allow the MDD to be fully
reduced. Specifically it does not take long edges or the possibility
that the root is not placed in layer $1$ into consideration. If the
root is in a layer $i > 1$, we will simply add a long edge that skips
layer $1$ to $i-1$, such that we only need to deal with long edges.

\subsection{Handling long edges} \label{subsec:long_edges} For the
purpose of valid domains computation, the important observation is
that if a valid long edge exists skipping layer $i$ to $j$, then all
available domain values for the corresponding variables are also
supported. For static valid domains computation \cite{VD-NOTE} these
variables are found by for each level finding the longest outgoing
edge. Given this information it is simple to list the variables which
have full support due to a long edge skipping over their layer in time
$O(n)$. We obtain a dynamic equivalent of this algorithm by
maintaining the length of the longest long edges originating from each
level. Additionally each long edge $e$ is placed in the support list
of $(x_{s(e)},v(e))$, as if it was a regular edge. In order to track
the longest outgoing edge of each level we will maintain the set $L$
of distinct intervals $[i,j]$ such that there exists at least one long
edge skipping layers $i$ to $j$. In order to detect when a skipped
interval should be removed from $L$, we for each skipped interval
$[i,j] \in L$ maintain a counter $l_{i,j}$, giving the number of long
edges skipping exactly the layers $i$ to $j$. A long edge $e$ skipping
layer $i$ to $j$ can no longer support the skipping of the interval
$[i,j]$ iff $e$ is invalidated. In this case $l_{i,j}$ is decremented,
and in case $l_{i,j} = 0$ the skipped interval $[i,j]$ is removed from
$L$. Note that we do not perform any update if a \proc{Remove} call
actually 'cuts' a long edge into two parts. This means that we allow
the intervals to support values that are invalid. However, the only
invalid values supported in this way are those that are given as
arguments to \proc{Remove}, and have therefore already been removed
from the valid domain values by another constraint.

Recall that our goal is to maintain the longest outgoing edge from
each layer in the MDD. Using the above we can do this simply by
maintaining a max-priority queue for each level $i$ over the the set
of skipped intervals of the form $[i,j]$, for some $j$, using $j$ as
key. Assuming the priority queue supports reporting the maximum in
time $O(1)$ we can obtain the required table in $O(n)$ time. It should
now also be clear why we build the priority queue over the distinct
intervals instead of all the intervals represented by long edges. Had
we used the long edges directly, all edge deletions would require a
delete operation on the priority queue, while we now only need a
delete operation when a distinct interval no longer has any supporting
long edges. The algorithm is given in Figure \ref{code:inc_remove} and
the following lemma gives its complexity.

\begin{figure} 
\begin{codebox}
\Procname{$\proc{Init}()$}
\li Initialize each support list $s_{i,v}$
\li $L \gets$ the set of intervals skipped by at least one long edge
\li $l_{i,j} \gets $ the number of long edges skipping exactly layer $i$ to $j$
\li $D^s_i \gets$ the values for $x_i$ supported by the support lists
\li $U \gets \{ 1, \ldots, n \}$ 
\li \proc{Remove}$(\emptyset)$ \RComment Initial domain propagation
\end{codebox}
\begin{codebox}
\Procname{$\proc{Remove}(R)$}
\li \For each $(x_i,v) \in R$
\li \Do \For each $(u,c) \in s_{i,v} $ \RComment{Remove edges supporting $(x_i,v)$}
\li $\proc{RemoveEdge}(u,c,v)$
\End
\li $U' \gets$ union of the intervals in $L$
\li \For each $l \in U \setminus U'$ \RComment{Layer $l$ is no longer skipped by a long edge}
\li \Do $D_l = D^s_l$
\li \If $D_l = \emptyset$ 
\li \Then Constraint failed
\End
\End 
\li $U \gets U'$
\end{codebox}
\begin{codebox}
\Procname{$\proc{RemoveEdge}(u,c,v)$}
\li $s_{l(u),v} \gets s_{l(u),v} \setminus \{(u,c)\} $
\li \If $s_{l(u),v} = \emptyset$ \RComment{$(x_{l(u)},v)$ no longer supported in support lists}
\li \Then $D^s_{l(u)} \gets D^s_{l(u)} \setminus \{v\}$
\End

\li $C_u \gets C_u \setminus \{ (c,v) \}$ 
\li $P_c \gets P_c \setminus \{ (u,v) \}$ 
\li \If $l(c) - l(u) > 1$ \RComment{If the removed edge is a long edge} 
\li \Then decrement $l_{l(u)+1,l(c) - 1}$
\li \If $l_{l(u)+1,l(c) - 1} = 0$
\li \Then $L \gets L \setminus \{ [l(u)+1,l(c)-1] \}$  
\End 
\End
\li \If $C_u = \emptyset$ \RComment{$u$ has no outgoing edges }
\li \Then \For each $(p,v') \in P_u$ \RComment{Remove incoming edges}
\li \Do $\proc{RemoveEdge}(p,u,v')$
\End
\End

\li \If $P_c = \emptyset$ \RComment{$u$ has no incoming edges }
\li \Then \For each $(c',v') \in C_c$ \RComment{Remove outgoing edges}
\li \Do \proc{RemoveEdge}$(c,c',v')$ 
\End
\End
\end{codebox}
\caption{\proc{Remove} calls \proc{RemoveEdge} for each edge that must
  be removed due the restrictions in $R$. Afterwards it recomputes the
  layers supported by long edges $L$ and updates the domains
  accordingly. \proc{RemoveEdge} takes as input an invalid edge in
  form of the source node $u$, the destination node $c$ and the
  corresponding value label $v$. It then moves from the invalid edge
  downwards in depth first manner as long as there are nodes being
  invalidated due to lacking valid incoming edges. If the node $u$ has
  no more outgoing edges it propagates upwards, removing nodes that
  have no more outgoing edges. Recall that $P_u$ corresponds to the
  incoming edges of $u$ and $C_u$ to the outgoing edges. For each edge
  it removes it decrements the counter of the corresponding skipped
  interval and updates $L$ if needed.}
\label{code:inc_remove}
\end{figure}

\begin{lemma}
On a root-to-leaf in the search tree the complexity of the
longest-outgoing-edge based approach is bounded by $O(|E| + n^2 \lg\lg
n + n^2d_{max})$.
\end{lemma}
\begin{proof}
  As previously, the actual time for handling normal edges is at most
  $O(|E|)$. Each interval can only be removed once, and each such
  deletion costs $O(\lg\lg |L|)$ using a VEB-based priority queue
  \cite{VEB}. Finally we spent $O(n)$ time per step to compute the
  table of longest outgoing edges and compute the variables covered by
  long edges. In total this yields a complexity over a root-to-leaf
  path in the search tree of $O(|E| + |L|\lg\lg |L| + tn)$, where $t$
  is the number of steps. Since there can at most be $O(n^2)$ distinct
  long edge intervals and $nd_{max}$ steps this yields $O(|E| + n^2
  \lg\lg n + n^2d_{max})$.
\end{proof}

We believe that for most practical applications of the MDD constraint
the above complexity will be completely dominated by the $|E|$ factor,
and hence that the addition of long edges result in no significant
performance impact. As an alternative solution we can use the dynamic
interval union data structure (DIU) presented in
\cite{DYN-INTERVAL-UNION} to store the intervals. Using this data
structure it is possible to obtain a complexity of $O(|E| + n^2\lg n +
nd_{max})$ \cite{MDD-SEARCH-REPORT}. However for practical
applications, the longest-outgoing-edge approach using a simple binary
heap is most likely preferable (yielding $O(|E| + n^2\lg n +
n^2d_{max})$) due to the simple implementation and low overhead.

\section{Maintaining the reducedness property} \label{sec:reduction} In the
above we do not take steps to maintain the uniqueness reduced property
of the MDD when we update the data structure. This forfeits a chance
for a large speed-up. If a reduction at an early search node $s$ would
lead to large reduction in the size of the data structure all
descendant search node of $s$ (of which there can be an exponential
number) would benefit from working on a much smaller data
structure. As an example of the effect of dynamic reduction consider
the simple constraint $f$ encoding the rule $x_1 \le x_2, x_1 \le x_3
\ldots, x_1 \le x_j$ with domains $D_i = \{1, \ldots, k \}$ for some
constant $k$. Let $MDD(f)$ be the MDD representing $f$ and let $f_v =
f \land x_1 = v$.  Now consider the removal of the value $1$ from the
domain $x_2,\ldots, x_j$ (as could be induced by an external
AllDifferent constraint). With this restriction $MDD(f_1)$ becomes
equivalent to $MDD(f_2)$ and can be merged, reducing the size of the
MDD required to store the constraint with a constant factor. If the
value $2$ is lost next then a further constant fraction of the MDD can
be removed due to the reduce step as $f_1 = f_2$ now becomes
equivalent to $f_3$. This is of course a very simplistic constraint
easily propagated using other methods, but if we consider the
conjunction of the constraint with another constraint, the example
still applies in many cases, especially if the new constraint does not
depend on the value of $x_1$. One example of such an additional
constraint is $\forall i \in [2,j-1] : x_i \not = x_{i+1}$. Note that
if we ensure the uniqueness reduced property the MDD will be fully
reduced according to the original domains throughout the search
(assuming it is fully reduced initially), since there is no risk that
new long edges will be created when we only perform domain
restrictions. We therefore first discuss how to ensure the uniqueness
reduced property and then describe the addition necessary in order to
obtain full reduction according to the \emph{current} domains.

\subsection{Dynamic reduction}
Assuming that the MDD is uniqueness reduced for all layers later than
layer $i$, a node $u$ in layer $i$ becomes redundant iff at least one
of its outgoing edges are modified such that $C_u = C_q$ for some node
$q$ in layer $i$. This redundancy can be removed by merging the
redundant nodes. Two redundant nodes $u$, $q$ are merged by removing
$u$ and redirecting all edges ending in $u$ to $q$ or vice versa. In
the first case we say that $u$ is the subsumee, and $q$ the
subsumer. Note that a merger can give rise to redundancies in earlier
layers, but not in later layers. We will therefore dynamically reduce
the MDD by maintaining maintain a set of 'dirty' nodes that need to be
checked for redundancy during the operation of
$\proc{Remove}$. Afterwards we check the dirty nodes for redundancy in
a bottom-up manner, ensuring that later layers are uniqueness reduced,
before earlier layers are considered.

\subsubsection{Redundancy detection}
In order to efficiently check whether a given node has become
redundant we will for each node maintain a hash signature computed
based on the node's level and outgoing edges. Using this signature and
a hashtable we can discover nodes that become redundant. Some care
must be take to ensure that the signature can be updated in $O(1)$
time when an edge changes. This can be ensured using a variation of
vector hashing \cite{VECTOR-HASHING}, which we describe in
\cite{MDD-SEARCH-REPORT}. We would like to stress that vector hashing
has a very low overhead, and has been shown to be competitive in
practice. Additionally, to ensure that collisions in the hash table do
not give rise to expensive comparisons between the outgoing edges of
two nodes, we utilize the technique of generating a hash signature
longer than necessary to index the hash table. Using this technique
nodes need only to be compared when their full signature matches, even
though they reside in the same bucket in the hash table. In this way
it is possible to ensure that inserting a node which is not redundant
takes expected time $O(1)$ and that insertion of a redundant node $u$
takes expected time $O(|C_u|)$. A detailed description can be found in
\cite{MDD-SEARCH-REPORT}.

\subsubsection{Merging nodes}
Given two nodes $u_1$ and $u_2$ to merge we always designate the one
with the largest number of parents as the subsumer in order to reduce
the total cost of the merge operations. An edge $e$ is only redirected
when its end-point $c$ is subsumed. Since this only happens when
another node of larger in-degree becomes identical to $c$ the
in-degree required to cause $e$ to be redirected must at least double
each time $e$ is moved. Hence an edge can only be moved $\lceil
\lg(|V|) \rceil$ times as $|V|$ is an upper bound on the in-degree of
a node.  Note that this is a very simple and classic greedy strategy
that incurs no significant overhead.

\subsubsection{Complexity of dynamic reduction}
\begin{lemma}
  The expected time spent over a root-to-leaf path in the search tree
  by \proc{Remove} on reducing the MDD is $O(|E|\lg |V|)$.
\end{lemma}
\begin{proof}
  Since each edge can only be redirected $\lceil \lg(|V|) \rceil$
  times the total cost of redirecting edges is
  $O(|E|\lg|V|)$. Checking and discovering a redundant node $u$ and
  removing its outgoing edges takes time $O(|C_u|)$. Note that each
  end-point of the deleted edges must have an in-degree of at least
  two before the merger, and therefore no further edges will need to
  be removed. This takes time at most $O(|E|)$ as edges cannot be more
  than once over a root-to-leaf path in the search. Checking a node
  that is not redundant takes expected time $O(1)$, and occurs only
  once per redirection or removal of an edge. The total expected time
  for this is therefore $O(|E|\lg |V|)$, by the bound on the number of
  edge redirections.
\end{proof}

\subsection{Full reduction based on current domains}\label{subsec:full_reduce}
The reduce step described above keeps the MDD fully reduced according
to the original domains. This means that while the MDD is uniqueness
reduced it is not fully reduced according to the \emph{current}
domains. As an example consider an MDD with 1 variable $x_1$ and a
single node $u_1$ with edges labelled $1$ and $2$ going to
\textit{terminal}. If the domain of $x_1$ is $\{ 1,2,3 \}$ this MDD is
fully reduced, while it reduces to the \textit{terminal} node if the
domain is $\{1,2\}$. Let $v(u)$ to denote the set of values supported
by $u$, then full reduction according to the current domains can be
achieved by applying the following rule: If there exists a node $u$
for which $v(u) = D_{l(u)}$ such that all outgoing edges from $u$ have
the same endpoint we will consider it redundant and reduce it into a
long edge. We observe that it is now possible for a node $u$ to become
redundant without having its outgoing edges modified: Consider a node
$u$ in layer $i$ such that all its outgoing edges lead to the same
node and and $v(u)=D_i \setminus \{ v \}$, where $v \in D_i$. Should
$v$ be removed from $D_i$ through a restriction, $u$ will now be
redundant. In order to detect this efficiently we will keep track of
the set of nodes $V_R = \{u \in V \mid \exists c \in V : \forall
(c',v) \in C_u : c' = c \}$, as these nodes are the only candidates
for being reduced using the new reduction rule. Since $u$ must either
be in $V_R$ initially or enter $V_R$ as children of $u$ are merged, we
can maintain $V_R$ during merge operations. This is easily done in
$O(1)$ per edge modification and thus do not affect the asymptotic
complexity.

When a domain is modified we need to find and reduce all nodes $u \in
V_R$ for which $v(u) = D_i$. This can be done efficiently by creating
a hash table that maps nodes in $V_R$ using $v(u)$ as key. Should the
domain $D_i$ be modified, reducible nodes can be found simply by
looking up $D_i$ in the hash table. This lookup yields all the nodes
of $V_R$ for which $v(u) = D_i$, which are exactly those nodes for
which the new reduction rule applies. The hash signature of the nodes
and domains can be maintained in $O(1)$ per edge or domain
modification using vector hashing. The space used for this is
insignificant compared to the space required to store the edges of the
MDD since it is only a small subset of $V$ that is inserted into the
table. Furthermore, since a node can only be removed by a reduction
once over a root-to-leaf path in the search tree this does not affect
the amortized complexity of the previous reduction technique.

\subsection{Domain entailment  detection} \label{subsec:domain_entailment} 
Given a constraint $F_k$, and a partial assignment $\rho$, let
$sol_\rho(F_k)$ be the set of vectors of domain values corresponding
to solutions allowed by $F_k$ that are consistent with $\rho$. A
constraint $F_k$ is said to be \emph{domain entailed} under domains
$D$ iff $\times_{1 \le i \le n} D_i \subseteq
sol_{\rho}(F_k)$\cite{vanhentenryck94design}. That is, if all possible
solutions to the CSP based on the available domains will be accepted
by $F_k$, then $F_k$ is entailed by the constraints implicit in the
domains. It is beneficial to be able to detect domain entailment as it
allows the solver to disregard the entailed constraint until it
backtracks through the search node where the constraint was first
entailed. If an MDD is kept fully reduced according to the current
domains it is entirely trivial to detect domain entailment as an
domain entailed MDD will be reduced to the \textit{terminal} node. If
the MDD is only kept uniqueness reduced and is domain entailed it is
easy to see that it will consist of a path of up to $n+1$ nodes. Note
that this state of the MDD is both necessary and sufficient for domain
entailment assuming that the MDD has performed the most recent domain
propagation step. Hence, checking for domain entailment just require
that we maintain a node count for each level. If all node counts are
one or less, and the constraint has not failed, the constraint is
domain entailed.

\section{Constructing the MDD}\label{sec:construction}
In order to apply our approach we first need to construct the MDD and
compute the necessary auxiliary data structures. The input to
constructing the MDD is assumed to be a set of constraints expressed
in discrete variable logic. For example, tabular constraints could be
expressed as disjunction of tuples, while an AllDifferent could be
expressed as $\forall (x_i,x_j) \in X^2, i \not = j : x_i \not =
x_j$. We suggest to construct the MDD by first building the ROBDD of
the component constraints using $\lg d_i$ binary variables to
represent domain values for $x_i$ (see for example
\cite{BDD-CONFIG}). This allows utilization of the optimized ROBDD
libraries available. Assuming that the binary variables encoding each
domain variable are kept consecutive in the variable ordering it is
trivial to construct the MDD from the BDD using time linear in the
resulting MDD. This also means that the many variable ordering
heuristics available for BDDs, which can substantially reduce the size
of the BDD, can be utilized as long as they are restricted to grouping
the binary variables according to the corresponding domain
variable. The initial data structures required can be computed using a
simple DFS scan of the MDD. The time that is acceptable for the
compilation phase (and therefore also the allowable size of
intermediate and final MDDs) depend on whether the constraint system
is to be solved once or whether it is used in for example a
configurator where the solver is used repeatedly on the same
constraint set (with different user assignments) to compute the valid
domains\cite{CONFIG-COMPARE}. One could easily specify a large set of
constraints and incrementally combine them into fewer and fewer MDD
constraints until a time or memory limit is reached and still gain the
benefit of improved propagation.

\section{Application to other data structures}\label{sec:applications}
The techniques covered in this paper can also be applied to some other
data structures that represent a compilation of a solution space. In
this section we briefly outline what can be achieved for two specific
data structures which are very similar to ROMDDs.

\subsection{AND/OR Multi-Valued Decision Diagrams (AOMDDs)}
AOMDDs were introduced in \cite{AOMDD}, and from the perspective of a
GAC algorithm can be thought of as introducing AND nodes into the MDD,
such that each child of an AND node roots an AOMDD which scope is
disjoint from its siblings. This data structure is potentially more
compact than an MDD. We note that the support list technique for
maintaining the valid domains is also applicable to AOMDDs. The only
change is that an AND node becomes invalid if it loses any of its
outgoing edges as opposed to all outgoing edges for an OR
node. Furthermore since an AOMDD is reduced in a manner similar to an
ROBDD our approach for dealing with long edges and obtaining dynamic
reduction can also be applied without substantial changes.

\subsection{Interval edges}\label{subsec:interval-edges}
In an ordinary MDD each edge corresponds to a single domain value,
except for long edges which in skipped layers represent the entire
domain. It is quite natural to consider the generalization to edges
that represent a subset of a domain. One particular useful
generalization of edges is to let each edge correspond to some
interval of the domain values. This approach is used in Case
DAGs \cite{SICTUS-PROLOG-MANUAL,BCC} which resemble MDDs without long
edges, but where edges represent disjoint intervals instead of single
values. If we use the technique from Section \ref{sec:simple-VD} and
construct support lists we will use an excessive amount of space and
no longer be able to claim a $O(|E|)$ complexity over a root-to-leaf
search path as each edge can occur in up to $d_{max}$ support
lists. This can be avoided by abandoning support lists and instead
observing that interval edges in many ways are similar to long
edges. By keeping track of the distinct intervals of values supported
by at least one edge in each layer, computing the valid domains
reduces to computing the union of this set of intervals. Based on this
observation it is possible to achieve a complexity of $O(nd_{max}^3 +
|E|)$ for maintaining GAC over a root-to-leaf path in the search
tree. The details can be found in \cite{MDD-SEARCH-REPORT}. This
compares with a worst-case complexity of $\Omega(d_{max}|E|)$ for a
\emph{single} step using a naive DFS approach. Additionally our
dynamic reduction technique is also applicable, the only non-trivial
change being that interval edges must be updated if their beginning or
end values are removed from the current domains. If we fail to perform
this update an edge $e=(u,c)$ for which $l(e)=[v_1 + 1,v_2]$ will not
be considered identical to an edge $e' = (u,c)$ for which
$l(e)=[v_1,v_2]$ when $v_1$ is removed from the domain of $x_{l(u)}$.

\section{Conclusion}
This paper introduced the ROMDD global constraint and thereby provided
a possible solution to the problem of representing ad-hoc constraints
in constraints satisfaction problems. In addition to providing an
efficient GAC algorithm for this constraint, we have shown how to keep
the underlying decision diagram reduced dynamically during search in
an efficient manner that also allows efficient domain entailment
detection. The MDD global constraint can be used to efficiently
represent the solution space of a set of simpler constraints. Based on
whether the constraint problem needs to be solved once or many times,
the time allocated to including smaller constraints in the MDD
constraint can be easily adjusted. Since the constraint uses a reduced
decision diagram to represent the solution space of the constraint it
can also be used to represent tabular constraints in a compressed
manner while still allowing a complexity that relates to the size of
the data structure and not the number of solutions stored. Finally we
have demonstrated how our approach can be applied to the AOMDD and
Case DAG data structures.

\bibliographystyle{splncs}
\bibliography{bib}
\end{document}